\documentclass[acmtog,nonacm,balance=false]{acmart}

\setcopyright{none}

\usepackage{graphicx}
\usepackage{cleveref}
\usepackage{multirow}
\usepackage{rotating}
\usepackage{wrapfig}

\usepackage{xcolor}

\usepackage{algorithm}
\usepackage{algorithmic}

\usepackage{soul}

\definecolor{myblue}{RGB}{40,80,220}
\definecolor{myorange}{RGB}{226,124,38}
\definecolor{myfuchsia}{RGB}{230,50,180}

\newcommand{\edit}[1]{#1}

\acmJournal{TOG}
\begin{document}
% Title portion

\title{Latent Space Jam: Layout Guided Video Synthesis}


\author{Asaf Joseph}
\authornote{Both authors contributed equally to this research.}
\orcid{0009-0000-5144-0602}
\email{asaf.joseph@mail.huji.ac.il}
\affiliation{%
  \institution{Lightricks}
  \city{Jerusalem}
  \country{Israel}
}

\author{Moshe Mandel}
\authornotemark[1]
\orcid{0009-0004-4240-8095}
\email{mandelmoshe@gmail.com}
\affiliation{%
  \institution{Independent}
  \city{Jerusalem}
  \country{Israel}
}

\author{Ohad Fried}
\orcid{0000-0001-7109-4006}
\affiliation{%
  \institution{CANVAS Lab, Reichman University}
  \city{Herzliya}
  \country{Israel}
}

\renewcommand{\shortauthors}{Joseph et al.}

\begin{abstract}
Adding control to video generation often requires training, which demands additional computational resources, data, or architectural changes. We propose a training-free approach that utilizes pretrained video diffusion models to control object layout and motion using coarse layouts. Our method operates in two passes: first, it steers spatiotemporal placement through prompt-guided cross-attention to produce a coarse visual guide; second, this guide conditions the same model to generate a high-quality, layout-consistent video. The approach enables coordinated motion with strong temporal consistency, supporting diverse trajectories and object interactions without additional training.
Extensive experiments show that our method consistently outperforms prior work in overlap and trajectory accuracy, as well as visual quality across static, dynamic, and multi-object scene settings.
\end{abstract}




% \centering
% \rule{5cm}{3cm}
% \caption{Test caption}

\begin{teaserfigure}
\centering
% One file so arXiv HTML keeps the layout and frames in a single row.
\includegraphics[width=\textwidth]{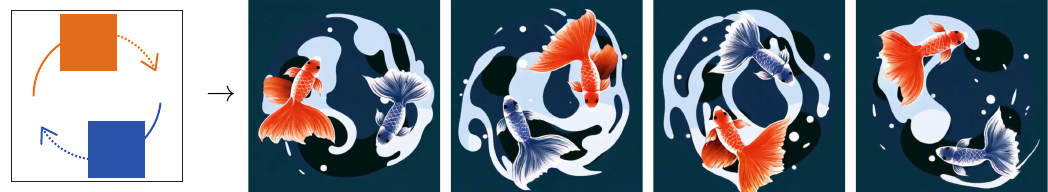}
\Description{Guiding a video using layout and text.}
\caption{We steer the output (right) of a pre-trained video diffusion model towards a given layout (left) without training.}
\end{teaserfigure}

\maketitle
% Running heads ("Joseph et al.") come from acmart's page style.
\pagestyle{plain}

\noindent\textbf{Project page:} \url{https://latent-space-jam.github.io/}

\section{Introduction}

Frame composition lies at the core of visual storytelling, defining the spatial structure through which meaning, motion, and narrative coherence emerge in video.
Recent advances in text-based video generation \cite{hacohen2026ltx, wan2025} have demonstrated a remarkable ability to generate diverse and creative results from free-form text prompts. Yet the absence of explicit mechanisms for controlling spatial composition within and across frames continues to limit their applicability in structured visual storytelling.

We consider the problem of layout-guided text-to-video generation. Given a pre-trained off-the-shelf video generator, our goal is to steer its output to match a desired layout.  Our method, Latent Space Justified Attention Mapping (JAM), allows the user to align the output video to a desired layout, guided by bounding boxes across frames, without additional training or fine-tuning.

Training-free layout-guided video generation has been explored in prior works \cite{jain2024peekaboo, lian2023llm, trailblazer, qiu2024freetraj, huang2025imzero}.
\edit{However, existing methods fail in scenarios involving complex motion, exhibiting limited dynamics and poor temporal coherence.}
In contrast, our approach is able to synthesize complex and dynamic motion while preserving the high visual fidelity of the pre-trained video diffusion backbones, without introducing additional training or degrading their generative capacity. Furthermore, both subjective human evaluations and objective quantitative metrics demonstrate that our method consistently outperforms these prior approaches (see \nameref{sec:results}).

There is an inherent trade-off between layout adherence and visual quality: stronger layout regularization results in poorer visual quality. Instead of trying to achieve both goals simultaneously, and thus compromising on both, we take a different route.
We split the pipeline into a guide generation stage followed by a refinement stage. Our key insight is that this allows us to push layout adherence to the maximum in the first stage, and rely on the robustness of modern video diffusion models in the refinement stage to fix the coarse intermediate results, thus circumventing the tradeoff between layout adherence and visual quality.

\edit{Inspired by attention-based control methods in text-to-image generation \cite{attend_and_excite, xie2023boxdiff}, we manipulate the cross-attention maps of pre-trained video diffusion models to generate coarse guidance. However, extending these approaches to video is not straightforward: the added temporal dimension makes it more difficult to control object placement consistently across frames, and the scale of video DiT models makes optimization across all layers impractical. In contrast, we show that effective control can be achieved through selective intervention, modifying only a small subset of cross-attention layers and a limited number of diffusion steps. As a result, our method can be executed on a single GPU, with minimal time overhead relative to the inference time of the underlying backbone models.}

In practice, we treat a pre-trained diffusion transformer (DiT) \cite{peebles2023dit} video generator as a fixed computational graph and optimize its latent input rather than its weights. Given a textual prompt and a spatiotemporal layout, we first steer an intermediate video toward the desired structure by optimizing this latent variable, using a loss applied to selected cross-attention layers of the DiT to align them with the provided bounding boxes. Then, we rerun the generator and inject the resulting video at an early diffusion timestep, using it as a coarse prior that the model refines into the final result.

\noindent
Our contributions are summarized as follows:
\begin{itemize}
    \item We introduce a training-free framework for layout-guided text-to-video generation that enables explicit control over spatial composition and motion using simple bounding-box supervision.

    \item We present the importance of using a two-stage pipeline: utilizing layout guidance videos as priors to the final result. 
    
    \item We show that minimal intervention in a pre-trained video diffusion model via a subset of its cross-attention layers is sufficient to steer generation.
    
    \item We validate our approach through extensive experiments, demonstrating superior performance over prior methods across complex, dynamic, and multi-object scenarios.
\end{itemize}

% \moshe{emphasize novelty of optimization on video models, and novelty of the 2-stage paradigm.}

% \moshe{emphasize practicality: by running on low resolution, quantized models, and partial layers optimization, we can run on consumer GPU without sacrificing quality of results (see Results table below)}
\section{Related Work}

% \moshe{emphasize the novelty compared to image-based attention manipulation: image models are smaller and therefore optimization methods are applied on all layers, whereas we show that even partial optimization works. images have no temporal dimension, compared to temporal-only control (e.g. tempoControl) we control spatial AND temporal control in a contiuous manner.}

% \moshe{emphasize the novelty of using the 2-stage paradigm? how?}

% \moshe{Why are we different from Direct-a-video and from BoxDiff (image based optimization we are based on)?}

% \moshe{cite tempocontrol, switchcraft}

Our work falls within the line of research on training-free controllability for diffusion models, which exploits internal attention representations as test-time control signals. We address the problem of spatiotemporal layout and motion control in video generation, achieving structured control without training, fine-tuning, or architectural modification.

A growing body of work in text-to-image synthesis demonstrates that cross-attention can be used as an interface for test-time control. Several methods \cite{attend_and_excite, epstein2023diffusion} interpret attention as a form of semantic grounding and introduce latent optimization or self-guidance strategies to enforce token presence and object saliency during sampling. Others \cite{Chen_2024_WACV, xie2023boxdiff, Phung_2024_CVPR} reshape attention maps to align text tokens with user specified spatial layouts or regions, enabling box and region constrained generation. Additional approaches \cite{agarwal2023star, Cao_2023_ICCV} focus on attention segregation, refocusing, and mutual self-attention control to improve compositionality and consistency. Collectively, these works establish that attention is not merely an internal mechanism, but a controllable structure that can be manipulated at inference time to impose semantic and spatial constraints without retraining the model.

In video generation, most controllable methods rely on additional training or fine-tuning. A common strategy is to augment pre-trained video diffusion models with external adapters or control branches conditioned on structural signals such as edges or depth \cite{hu2023videocontrolnet},  pose or trajectories \cite{Hu_2024_animateanyone, karras2023dreampose, wang2024motionctrl, yang2024_directavideo}, a combination of these signals \cite{wang2023videocomposer, jiang2025vace}, or camera parameters \cite{bai2025recammaster, yang2024_directavideo}. These approaches enable explicit control over motion, layout, and viewpoint. However, they require paired supervision, dedicated retraining, and model modification, which restricts their applicability and adds significant computational cost.

Recently, several works have explored training-free control for text-to-video diffusion models.
Peekaboo \cite{jain2024peekaboo} and TrailBlazer \cite{trailblazer} intervene in the cross-attention layers of a pretrained model to align foreground motion with user-specified guidance.
FreeTraj \cite{qiu2024freetraj} adopts a similar strategy but further improves controllability by imposing guidance on both attention maps and the latent noise.
LVD~\cite{lian2023llm} focuses on generating a layout guide with an LLM, producing bounding box trajectories across scenes to steer object placement.
\edit{SwitchCraft \cite{xu2026switchcraft} and TempoControl \cite{TempoControl2025} apply a training-free approach to modify contemporary DiT video models, but focus only on temporal control. }

In contrast to prior training-free video control methods that rely on direct attention and latent noise manipulation \edit{\cite{jain2024peekaboo,trailblazer, qiu2024freetraj}}, our approach adopts an optimization-based attention alignment strategy that scales effectively to complex spatiotemporal layouts. This enables stronger layout adherence while preserving motion diversity and visual fidelity, as confirmed by both quantitative metrics and human evaluations (see \nameref{sec:results}).

\section{Preliminaries}

\subsection*{Video Diffusion Models}

Diffusion models generate data by learning to reverse a gradual noising process. Given a clean sample $x_0$, a forward process progressively adds Gaussian noise over $T$ steps to obtain $x_T$. A neural network $\epsilon_\theta$ is trained to predict the noise at each timestep, enabling sampling by iterative denoising from pure noise.

Modern video diffusion models extend image diffusion to the spatio-temporal domain by operating on sequences of frames. Most state-of-the-art approaches adopt a latent diffusion formulation, where a variational autoencoder encodes frames into a compact latent representation $z$, and diffusion is performed in latent space rather than pixel space. 

Recent approaches replace convolutional U-Nets with Diffusion Transformers, which operate on spatio-temporal latent tokens using self-attention. This design better captures long-range spatial and temporal dependencies, improving temporal coherence and scalability. 
In this work, we assume a latent video diffusion model parameterized as
\begin{equation}
    \epsilon_\theta(z_t, c, t),
\end{equation}
where $z_t$ is the noisy latent at timestep $t$ and $c$ is a conditioning signal.

\subsection*{Test-Time Latent Optimization via Cross-Attention}

Text-conditioned diffusion models incorporate conditioning information through cross-attention layers that relate conditioning tokens (e.g., text) to visual latent tokens. The resulting attention maps capture how semantic concepts influence spatial regions during the denoising process.

Test-time latent optimization utilizes this structure to steer generation without modifying model parameters. Instead of updating the network weights, the latent variable is optimized during sampling with respect to a task-specific objective. Concretely, given a target constraint $B$ (e.g., a target layout map), a differentiable loss is defined over model activations (e.g., cross-attention maps), and gradients are backpropagated to the latent variable:
\begin{equation}
    z_t \leftarrow z_t - \eta \nabla_{z_t} 
    \mathcal{L}(A(z_t, c), B),
\end{equation}
where $A(\cdot)$ denotes cross-attention activations, $c$ is the conditioning input, and $\eta$ is the step size.

This formulation treats cross-attention as an interface through which constraints can influence generation, enabling controllable inference via latent updates while keeping model parameters fixed.

\begin{figure*}[!ht]
  \centering
  \includegraphics[width=\textwidth]{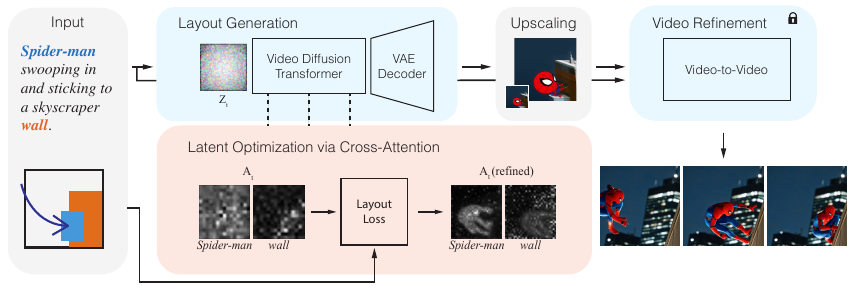}
  \caption{
  Overview of our training-free layout-guided video synthesis pipeline.
Given a text prompt and spatiotemporal layout constraints, we first perform Layout Generation, where the latents of a frozen text-to-video diffusion model are optimized using a cross-attention-based layout loss. This latent refinement process aligns token-specific attention with the target layout and produces a coarse spatiotemporal guiding video in latent space. Next, after the guide is optionally spatially and/or temporally upsampled, we render the final video: the guiding video is injected at an early timestep into the video diffusion pipeline. This second stage refines appearance and details while preserving the imposed layout and motion, yielding the final high-quality video.}
  % \Description{Diagram of layout guided video diffusion model pipeline}
  \label{fig:overview}
  \Description{A diagram of a training-free layout-guided video synthesis pipeline.}
\end{figure*}
\section{Latent Space Jam}

Enforcing layout adherence conflicts with the visual quality of a generated video.
In the early stage of generating the layout-guided video, more complex motion requires stronger layout regularization, thereby reducing visual fidelity.
Our main insight is that we can circumvent this tradeoff via a two-stage process. We first allow better layout adherence at the cost of visual quality, and then enhance this low-quality (but layout-accurate) guiding video. 

The enhancement is achieved using existing state-of-the-art video diffusion models \cite{wan2025, hacohen2026ltx},
which can be adapted to accept a video input, 
turning them into robust quality amplifiers that 
can correct visually poor videos while maintaining object locations and motion.

\Cref{fig:overview} provides an overview of our pipeline. We present a two-stage training-free layout-guided video synthesis framework that enables explicit control in both time and space over videos generated by a pre-trained video diffusion model, without any finetuning of model parameters. 
Given a text prompt and user-specified layout constraints in the form of bounding boxes across frames, we enforce the model to generate a guiding video that adheres to these layout constraints. Then, we use this guiding video as input to a video-to-video pipeline to generate the final video.

In the following, we motivate the need for a structural prior. Then, we analyze the limitations of explicit guide generation using ensembles of text-to-image models. Finally, we propose an alternative formulation in which the guide is generated implicitly by optimizing the latent input of a frozen text-to-video diffusion model.

\subsection{On The Importance of a Guiding Prior}

At the first stage of our pipeline, we generate a coarse guiding video prior, allowing us to focus fully on layout adherence. 

\Cref{fig:t2i-abl-figure} shows such a guiding video and its corresponding output, demonstrating the ability of existing video diffusion models to produce high quality video that preserves the intended layout from a low fidelity guide.
This shows that the guide itself does not need to be visually detailed, but should adhere as closely as possible to the intended layout.

Following the first stage, we adapt the second stage into a video-to-video pipeline by injecting the prior at an early diffusion time-step. At early time-steps, the diffusion process primarily governs the global layout, while later steps refine visual details \cite{wang2023outline_first}.
Specifically, similar to \cite{meng2022sdedit}, we inject a noisy version of the video prior at a selected time-step, matching its noise level to the step, and continue denoising to generate the final video with restored visual fidelity.
Technical details are provided in the supplemental.

This raises the question of how such a guiding video prior should be constructed. In particular, is it enough to provide unrelated frames with weak spatiotemporal coherence? We explore this question next.

\subsection{Are Text-to-Image Models Suitable for Guidance?}

A straightforward strategy to generate the low-fidelity layout prior is to leverage the strong semantic and compositional capabilities of recent text-to-image (T2I) models. Since the guide is only required to encode coarse spatial layout and approximate object motion, rather than high visual quality, it is natural to generate a sequence of images that satisfy the bounding-box constraints on a frame-by-frame basis and treat this sequence as a guiding video. This can be implemented by applying layout-guided and inpainting techniques \cite{li2023gligen, flux2024} to independently generate a static background image and the objects specified in the text prompt, and then placing and propagating these objects across frames according to the provided layout constraints.

As explained in \cref{sec:quant} and \cref{tab:quantitative_table}, this simple baseline (T2I) surprisingly matches or outperforms previous layout-guided video generation methods \cite{jain2024peekaboo, qiu2024freetraj, trailblazer} on refined videos, achieving comparable or better performance across all datasets in overlap accuracy (IoU, AP50) and trajectory accuracy (CD, I., I-AP50).

However, this baseline has fundamental limitations.
Each frame is generated independently, with objects created separately before being inserted into the scene. 
While this guarantees correct spatial placement, it does not preserve stable object structure or realistic interactions among objects and their environment. 
\Cref{fig:t2i-boy-scarf} shows a recurring failure mode of this method; further results are provided in the supplemental.

\subsection{Utilizing Video Models for Better Guidance}
\label{sec:optimizing}

Instead of relying on external text-to-image models to generate a prior, we leverage the video diffusion model itself to produce a layout-aware guidance signal. Specifically, we exploit the powerful generative capacity of video diffusion models to synthesize a guiding video that exhibits coherent motion while respecting specified layout constraints. 

To enforce spatial correspondence between the generated video and the target layout, we take inspiration from attention-based image editing methods~\cite{xie2023boxdiff, attend_and_excite}.
At a high level, we control the output of a video diffusion model by steering its cross-attention maps towards the provided layout. 

\begin{figure*}[!htbp]
\centering
\includegraphics[width=\textwidth]{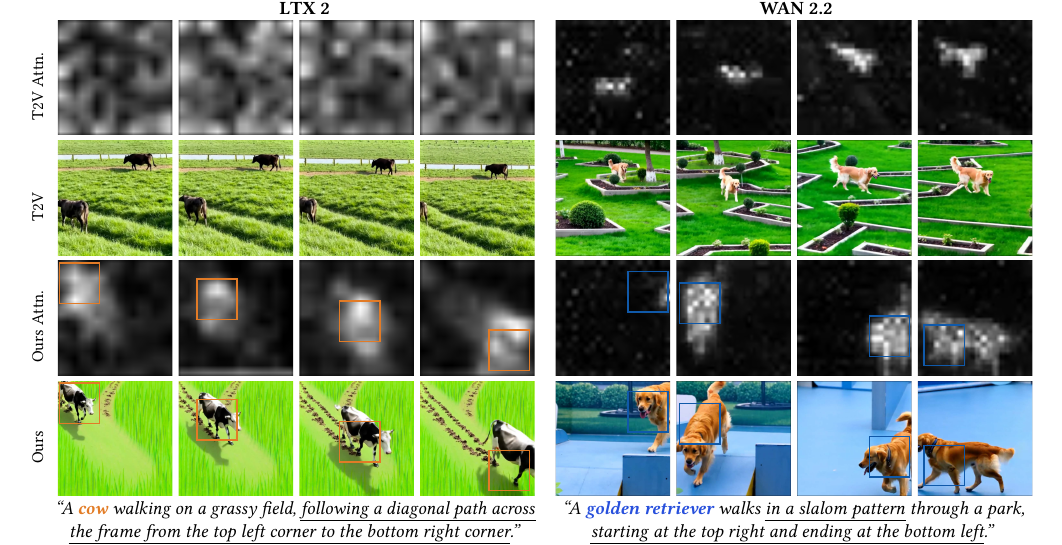}
\caption{
Outputs of Text-to-Video (T2V) and our method (Ours) across different backbone models (LTX-2, WAN 2.2), together with the corresponding attention maps extracted from an intermediate timestep. Attempting to control layout via textual prompt (underlined) fails, whereas controlling the attention map results in a video that corresponds to the desired layouts (orange/blue bounding boxes).
}
\label{fig:attn}
\Description{Comparing attention maps and their corresponding outputs.}
\end{figure*}

Given that the model’s intermediate cross-attention maps exhibit strong spatial correspondence with the final video, optimizing the latent inputs to minimize a loss between these maps and the target layout compels the generated output to follow the intended arrangement. By contrast, attempting to control the layout only via text fails.  We demonstrate this in \Cref{fig:attn}.

\begin{figure*}[!htbp]
\centering
\includegraphics[width=\textwidth]{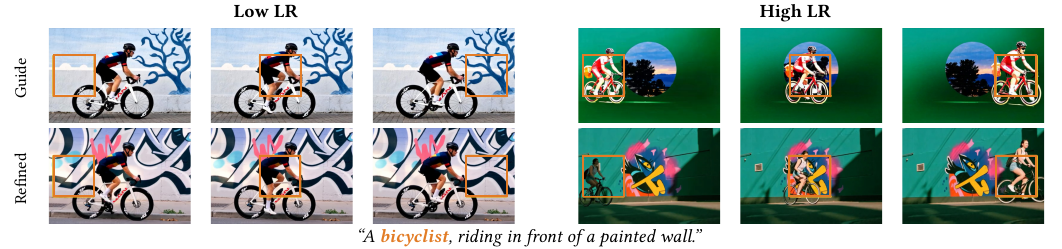}
\caption{
Comparison of different learning rates during optimization. In each section, guide (top) and refined (bottom) videos are shown.  The left column (low learning rate) produces guide videos with weaker adherence to the intended layout (indicated by orange squares). In contrast, the right column (high learning rate) achieves stronger layout adherence, but at the cost of reduced visual quality.
}
\label{fig:optimization-strength}
\Description{Comparison of generated videos of different learning rates, the videos show a person on a bicycle riding from left to right of frame.}
\end{figure*}

This optimization prioritizes structural alignment over visual quality. Stronger attention constraints enforce the target layout more accurately, but degrade appearance. As stated earlier, a coarse video is sufficient to guide generation when injected at early diffusion steps. This allows us to optimize aggressively for layout in this stage and rely on a second diffusion pass to restore visual fidelity. We therefore obtain a controllable tradeoff between structure and appearance, which we demonstrate in \Cref{fig:optimization-strength} and discuss further in the \nameref{sec:results} section. 

In detail, we optimize the diffusion latents using a cross-attention-based loss during the generation process, by minimizing the $KL$ divergence loss defined over the model’s cross-attention maps:

\begin{equation}
\mathcal{L}_{\text{layout}}
\;=\;
\sum_{i=1}^{N} KL\!\left(\mathbf{q}^{(i)} \,\|\, \mathbf{p}^{(i)}\right)
\end{equation}

\begin{align}
\mathbf q^{(i)} =
\bigl(q^{(i)}_j\bigr)_{j=1}^{|V|},
\qquad
q^{(i)}_j \;=\;
\frac{B^{(i)}_j}{\sum_k B^{(i)}_k},
\\
\mathbf p^{(i)} =
\bigl(p^{(i)}_j\bigr)_{j=1}^{|V|},
\qquad
p^{(i)}_j \;=\;
\frac{\sum_{t\in\mathcal T_i} A_{j,t}}
{\sum_k \sum_{t\in\mathcal T_i} A_{k,t}}.
\end{align}

where $q^{(i)}$ and $p^{(i)}$ are the normalized target layout distribution and the model's cross-attention distribution, respectively. $\{B^{(i)}\}_{i=1}^N$ is a set of $N$ bounding-box binary masks defined over the $|V|$ spatiotemporal locations of the input video across all frames. Each mask $B^{(i)}$ is associated with a set of textual tokens $\mathcal{T}_i$. Here, $B^{(i)}_j\in\{0,1\}$ denotes the value of mask $B^{(i)}$ at spatiotemporal location $j$, and $A_{j,t}$ denotes the cross-attention value between the $j$-th spatiotemporal location and the $t$-th text token, averaged across all heads and selected layers. Further techniques for aggregating $p^{(i)}_j$ across tokens per layout mask were tested and are discussed in the supplemental.

\begin{algorithm}[t]
\caption{Latent Optimization via Attention-Based Layout Loss}
\label{alg:refinement}
\begin{algorithmic}[1]

\STATE \textbf{Input:} prompt $\mathcal{P}$, latents $\mathbf{z}_t$, model $f_\theta$, $R$ steps, step size $\eta$, masks $\{\mathbf{B}_i\}_{i=1}^{N}$, token sets $\{\mathcal{T}_i\}_{i=1}^{N}$
\STATE \textbf{Output:} optimized latents $\mathbf{z}_t^\star$

\STATE $z^1 = z_t$

\FOR{$r = 1$ to $R$}
    \STATE $\mathbf{A}_t \leftarrow f_\theta(\mathbf{z}^r, \mathcal{P})$
    \STATE $\mathbf{A}_t \leftarrow \mathrm{get\_avg\_attention}(\mathbf{A}_t)$
    \STATE $\mathbf{A}_t \leftarrow \operatorname{Softmax}(\mathbf{A}_t)$
    \STATE $\mathcal{L} \leftarrow 0$
    
    \FOR{$i = 1$ to $N$}
        \STATE define $q^{(i)} = (q^{(i)}_j)_{j=1}^{|V|}$
        \STATE define $p^{(i)} = (p^{(i)}_j)_{j=1}^{|V|}$
        \STATE $\mathcal{L} \leftarrow \mathcal{L} + \mathrm{KL}\!\left(\mathbf{q}^{(i)} \,\|\, \mathbf{p}^{(i)}\right)$
    \ENDFOR

    \STATE compute gradient $\nabla_{\mathbf{z}^r} \mathcal{L}$
    \STATE $\mathbf{z}_t^\star \leftarrow \mathbf{z}^r - \eta \nabla_{\mathbf{z}^r}\mathcal{L}$
    \STATE $\mathbf{z}^r \leftarrow \mathbf{z}_t^\star$
\ENDFOR

\STATE \textbf{return} $\mathbf{z}_t^\star$

\end{algorithmic}
\end{algorithm}

A complete breakdown of an individual refinement step is provided in \Cref{alg:refinement}.  We apply multiple refinement steps at selected diffusion time steps.
\edit{Unlike image-domain UNet models, video DiT models are substantially larger, making full-layer optimization infeasible and motivating our approach of selectively optimizing a subset of layers.}
To improve quality while reducing computational cost, we carefully select \edit{the} subset of layers for the optimization process. See \Cref{tab:abl_layers} for further analysis. 

\subsubsection*{Reducing Inference Time}
Optimizing on latents can be a lengthy process, and the running time grows larger as the latent size increases. We can substantially reduce inference time with a minimal impact on guide quality by running the optimization stage at lower spatial and temporal resolutions (see \Cref{tab:abl_stage1_resolution}).

% FIGURE PAGE 1
\begin{figure*}[!ht]
\centering
\includegraphics[width=\textwidth]{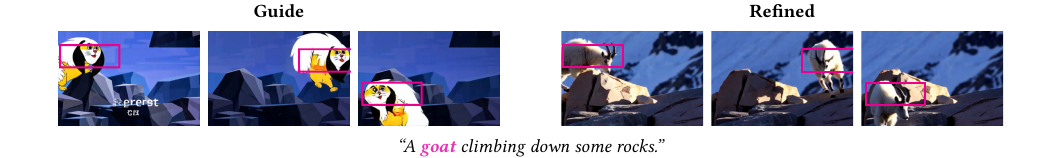}
\caption{
The guide (left) is of low visual quality but adheres to layout,
the refined video (right) improves the visual quality and maintains layout adherence.
}
\label{fig:t2i-abl-figure}
\Description{Comparison of two generated outputs of a coarse guide and a refined video. Both show a goat climbing down rocks in a slalom motion.}
\end{figure*}

\section{Results}
\label{sec:results}

\begin{table*}[!htbp]
\centering
\caption{Quantitative comparison with prior training-free layout-guided methods across datasets. IoU and AP50 measure overlap accuracy. CD, I., and I-AP50 measure trajectory accuracy. FVD and ClipSim measure visual quality. Our framework achieves superior spatial alignment and trajectory consistency on all metrics. Notably, Wan 2.2 excels in guide-stage performance, while LTX-2 provides comparable or stronger refined results.}
\setlength{\tabcolsep}{3.5pt}
\resizebox{\textwidth}{!}{
\begin{tabular}{l | c c c c c c c | c c c c c c c}
\toprule

& \multicolumn{7}{c|}{Guide}
& \multicolumn{7}{c}{Refined} \\
\cmidrule(lr){2-8}\cmidrule(lr){9-15}

& IoU ($\uparrow$) & AP50 ($\uparrow$) & CD ($\downarrow$) & I. ($\uparrow$) & I-AP50 ($\uparrow$) & FVD ($\downarrow$) & ClipSim ($\uparrow$)
& IoU ($\uparrow$) & AP50 ($\uparrow$) & CD ($\downarrow$) & I. ($\uparrow$) & I-AP50 ($\uparrow$) & FVD ($\downarrow$) & ClipSim ($\uparrow$) \\
\midrule
& \multicolumn{14}{c}{\textbf{STATIC}} \\
\midrule

PEEKABOO            & 17.02 &  8.80 & 24.74 & 45.17 & 45.10 & 139.00 & \textbf{30.98}
                   & 19.45 &  4.37 & 18.96 & 58.76 & 66.71 & 132.92 & 29.14 \\
FreeTraj            & 18.80 & 12.87 & 22.05 & 45.46 & 44.64 & 178.25 & 26.62
                   & 19.47 &  5.71 & 17.08 & 53.93 & 57.56 & 126.75 & 30.20 \\
TrailBlazer         & 20.32 & 12.40 & 20.65 & 46.98 & 45.97 & 151.23 & 30.30
                   & 16.95 &  5.37 & 19.42 & 44.65 & 45.13 & 126.76 & 30.26 \\
\edit{TrailBlazer (Wan 2.2)}   & 21.76 & 13.51 & 18.18 & 55.10 & 56.76 & 148.06 & 28.73
                   & 21.25 &  13.75 & 18.89 & 55.98 & 56.03 & 140.14 & 28.75 \\

T2I  & -- & -- & -- & -- & -- & -- & -- & 21.44 & 16.15 & 20.49 & 65.25 & 60.87 & 174.56 & 26.11
                     \\
% Ours (Wan 2.2)       & \textbf{43.78} & \textbf{42.11} & \textbf{10.84} & \textbf{82.29} & \textbf{84.92} & 145.95 & 27.04
                    % & 32.04 & 23.94 & 10.44 & \textbf{76.04} & \textbf{81.69} & 132.10 & 29.09 \\
Ours (Wan 2.2) & 29.51 & 15.90 & \textbf{12.47} & \textbf{75.05} & \textbf{84.01} & \textbf{118.25} & 30.86 & 
28.24 & 14.10 & 12.87 & \textbf{72.23} & 80.30 & \textbf{117.41} & 28.93 \\
Ours (LTX-2)         & \textbf{30.49} & \textbf{24.55} & 19.03 & 59.80 & 63.36 & 153.79 & 28.92
                    & \textbf{36.55} & \textbf{27.43} & \textbf{9.84} & 71.37 & \textbf{81.38} & 130.10 & \textbf{31.08} \\

\midrule
& \multicolumn{14}{c}{\textbf{DYNAMIC}} \\
\midrule

PEEKABOO            & 10.83 &  1.15 & 27.00 & 37.23 & 36.30 & 136.97 & \textbf{30.70}
                   & 11.89 &  1.15 & 25.06 & 42.72 & 43.45 & 126.75 & 29.37 \\
FreeTraj            & 17.39 &  5.26 & 20.49 & 46.80 & 41.60 & 140.85 & 30.62
                   & 13.91 &  1.44 & 21.68 & 44.09 & 42.41 & \textbf{110.90} & 29.94 \\
TrailBlazer         & 29.17 & \textbf{29.22} & 20.55 & 59.47 & 56.09 & 143.72 & 29.84
                   & 14.40 &  3.20 & 22.71 & 42.15 & 42.43 & 114.37 & 29.86 \\
\edit{TrailBlazer (Wan 2.2)} & 17.55 & 5.29 & 19.71 & 47.94 & 49.45 & 137.91 & 29.71
                   & 17.46 &  5.42 & 19.95 & 50.42 & 49.86 & 135.63 & 29.61 \\

T2I & -- & -- & -- & -- & -- & -- & -- & 27.23 & \bf{22.86} & 17.42 & 63.64 & 62.67 & 174.95 & 26.63
                     \\
% Ours (Wan 2.2)       & \textbf{45.95} & \textbf{44.79} & \textbf{8.06} & \textbf{88.05} & \textbf{91.67} & 165.76 & 27.08
                    % & 31.56 & 13.55 & \textbf{10.43} & 77.29 & 82.95 & 132.07 & 29.89 \\

Ours (Wan 2.2)      & 30.86 & 9.37 & \textbf{11.23} & \textbf{83.30} & \textbf{86.92} & 151.72 & 28.87
                    & 27.64 & 6.85 & 11.24 & \textbf{82.04} & 84.57 & 132.38 & 27.63 \\
                    
Ours (LTX-2)         & \textbf{33.70} & 20.90 & 16.28 & 70.28 & 74.70 & \textbf{128.85} & 30.14
                    & \textbf{35.24} & 18.79 & \textbf{10.74} & 79.76 & \textbf{88.42} & 119.30 & \textbf{31.13} \\

\midrule
& \multicolumn{14}{c}{\textbf{MULTI}} \\
\midrule

TrailBlazer & 8.41 & 7.65 & 42.72 & 31.97 & 26.51 & \edit{251.42} & 27.89 & 11.60 & 5.60 & 33.53 & 39.25 & 36.92 & \edit{235.19} & 30.00 \\ 
\edit{TrailBlazer (Wan 2.2)} & 16.46 & 14.65 & 31.83 & 44.61 & 35.87 & \edit{202.14} & \textbf{30.89} 
    & 13.86 & 11.01 & 32.47 & 38.41 & 33.51 & \edit{204.73} & 31.27 \\ 
T2I & -- & -- & -- & -- & -- & -- & -- & 11.95 & 8.88 & 36.11 & 35.89 & 34.27 & \edit{221.30} & 24.20 \\
    % Ours (LTX-2) & - & - & - & - & - \\ 
% Ours (Wan 2.2) & \textbf{31.52} & \textbf{30.92} & \textbf{24.26} & \textbf{63.04} & \textbf{67.38} & -- & 28.67 & 25.38 & 16.66 & 24.77 & \textbf{63.90} & \textbf{69.69} & -- & 31.08 \\

Ours (Wan 2.2) & \textbf{25.86} & \textbf{20.54} & \textbf{24.62} & \textbf{62.86} & \textbf{60.28} & \edit{\textbf{196.41}} & 30.60 & 25.83 & \textbf{20.73} & \textbf{21.87} & \textbf{64.57} & \textbf{69.03} & \edit{\textbf{198.57}} & 30.76 \\

Ours (LTX-2) & 21.88 & 20.07 & 29.30 & 52.96 & 50.35 & \edit{210.60} & 29.05 & \textbf{27.23} & {17.91} & {22.41} & 59.11 & 60.74 & \edit{228.52} & \textbf{32.05}  \\
    
\bottomrule
\end{tabular}
}
\label{tab:quantitative_table}
\end{table*}

% \moshe{we are already emphasizing that we are comparing only to training-free methods (which is why Direct-a-video is not compared to).}

In this section, we demonstrate the effectiveness of our method. The main focus of our technique it to generate videos according to the prompt and spatiotemporal layouts provided by the user. 

\edit{
We compare our method to the T2I-based method proposed above and to Peekaboo \cite{jain2024peekaboo}, TrailBlazer \cite{trailblazer}, and FreeTraj \cite{qiu2024freetraj}, prior training-free methods that generate videos using layout guidance.}

For \edit{fair} comparison of the guide generation stage, in addition to the provided benchmarks we use their outputs as guiding videos to the second stage of our method, and inject these outputs to a video diffusion model (see the refined section of \Cref{tab:quantitative_table}, right). 
\edit{Furthermore, we implement TrailBlazer \cite{trailblazer} on Wan 2.2, the backbone that we use in our method, as it achieved the strongest benchmark performance overall and is the only benchmark that supports multi-object settings.}

% , \edit{and implement TrailBlazer on Wan 2.2 as the underlying backbone, rather than its original U-Net–like model.}

\subsubsection{Datasets} To ensure a comprehensive evaluation across varying levels of difficulty, we consider three data categories: Static, Dynamic, and Multi. These contain approximately 40, 40, and 20 layouts, respectively, for a total of about 100 layouts.
The Static set contains single-object layouts with limited motion, initially based on videos from the Davis dataset \cite{davis_2017}.
The Dynamic set follows the same single-object structure with similar textual prompts as the Static set, but introduces strong dynamic motion.
The Multi set represents the most challenging scenario, featuring multiple objects with strong dynamic motion in a single video and custom textual prompts. 
For all datasets, layout guidance is provided in the form of binary video masks, which are created by the authors and to be publicly released alongside the textual prompts.
% \ohad{Where are the videos from in the Dynamic and Multi scenarios? How many are there? Where did we get the layout guidance from?}
% \moshe{added info.}

% \moshe{TODO: add FVD and ClipSIM}

\subsection{Quantitative Results}
\label{sec:quant}

% \sout{We compare our method to Peekaboo \cite{jain2024peekaboo}, TrailBlazer \cite{trailblazer}, and FreeTraj \cite{qiu2024freetraj}, training-free methods that generate videos using layout guidance, as well as our implementation of TrailBlazer on Wan 2.2 and the T2I-based method proposed above.}

We present results in \Cref{tab:quantitative_table}, additional results are in the supplemental.
To evaluate the control of the layout on the generated output, we use a set of complementary metrics that measure both semantic and spatial accuracy between the ground truth layout and bounding boxes predicted by an open-vocabulary segmentation model \cite{minderer2022owlvit} . Intersection over Union (IoU) measures the average overlap between predicted and ground-truth object masks across all objects and frames, providing an overall layout accuracy score. Average precision@50\% (AP50) evaluates frame-level correctness by considering a prediction correct when it overlaps the ground truth by at least 50\%, indicating whether objects appear in the intended regions per frame. To evaluate motion fidelity with respect to the target trajectory beyond standard overlap accuracy, we also report Centroid Distance (CD), Intersection, and Intersection-AP50 (I-AP50). CD measures how far the predicted object centers are from the ground-truth centers across all objects and frames. Intersection measures the raw overlapping area between predicted and ground-truth masks, and I-AP50 applies the detection-style metric based on the raw intersection area rather than IoU. Together, these metrics quantify how faithfully the generated videos follow the provided layouts. 
We employ Fr\'{e}chet Video Distance (FVD) \cite{frechet} to assess video quality against a set of 150 videos from Davis \cite{davis_2017}, and CLIP Similarity (ClipSim) \cite{pmlr-v139-radford21a-clipsim} to measure alignment with the textual prompts.

As shown in \Cref{tab:quantitative_table}, our method strongly outperforms other approaches in terms of alignment with the input and video quality across all datasets and metrics.
Wan 2.2 performs better in the guide stage, whereas LTX-2 excels after refinement, likely due to its dedicated refinement model. 
\edit{Notably, even when implemented on the same backbone (Wan 2.2) for a fair comparison, TrailBlazer continues to underperform our method across all stages, datasets and metrics.}

\subsection{Qualitative Results}
\begin{table*}[!htbp]
\centering
\caption{
User study results  \edit{(mean score $\pm$ confidence interval with 95\% confidence level)} across prompt adherence (Prompt), layout adherence (Layout), and visual quality (VQ). T2V excels in visual quality and prompt adherence but shows poor layout adherence. Our method achieves the strongest layout adherence while remaining competitive in prompt alignment and visual quality.
}
\resizebox{\textwidth}{!}{
\begin{tabular}{l|ccc|ccc|ccc}
\toprule
\multirow{2}{*}{Method}
& \multicolumn{3}{c|}{STATIC}
& \multicolumn{3}{c|}{DYNAMIC}
& \multicolumn{3}{c}{MULTI} \\
\cmidrule(lr){2-4}\cmidrule(lr){5-7}\cmidrule(lr){8-10}
& Prompt ($\uparrow$) & Layout ($\uparrow$) & VQ ($\uparrow$)
& Prompt ($\uparrow$) & Layout ($\uparrow$) & VQ ($\uparrow$)
& Prompt ($\uparrow$) & Layout ($\uparrow$) & VQ ($\uparrow$) \\
\midrule

T2V & \textbf{8.00 $\pm$ 1.76} & 6.50 $\pm$ 1.96 & 7.00 $\pm$ 1.24 & \textbf{9.25 $\pm$ 1.52} & 0.60 $\pm$ 1.11 & \textbf{7.75 $\pm$ 3.53} & 8.21 $\pm$ 0.90 & 3.62 $\pm$ 1.41 & \textbf{7.92 $\pm$ 1.06} \\

PEEKABOO & 5.00 $\pm$ 1.26 & 4.15 $\pm$ 1.13 & 3.58 $\pm$ 1.15 & 4.84 $\pm$ 1.34 & 1.19 $\pm$ 0.49 & 2.64 $\pm$ 0.92 & $N/A$ & $N/A$ & $N/A$ \\

TrailBlazer & 5.38 $\pm$ 1.36 & 4.15 $\pm$ 1.27 & 4.25 $\pm$ 1.37 & 1.60 $\pm$ 0.76 & 4.15 $\pm$ 1.59 & 1.48 $\pm$ 0.57 & 0.83 $\pm$ 0.75 & 1.39 $\pm$ 0.89 & 1.54 $\pm$ 0.92 \\

FreeTraj & 2.58 $\pm$ 0.95 & 3.77 $\pm$ 1.31 & 1.62 $\pm$ 0.55 & 2.92 $\pm$ 0.95 & 4.46 $\pm$ 1.37 & 1.96 $\pm$ 0.69 & $N/A$ & $N/A$ & $N/A$ \\

PEEKABOO [refined] & 7.33 $\pm$ 0.86 & 6.08 $\pm$ 1.14 & 7.25 $\pm$ 0.96 & 5.68 $\pm$ 1.49 & 1.39 $\pm$ 0.91 & 5.72 $\pm$ 1.06 & $N/A$ & $N/A$ & $N/A$ \\

TrailBlazer [refined] & 7.50 $\pm$ 1.01 & 5.00 $\pm$ 1.47 & 7.00 $\pm$ 0.98 & 7.24 $\pm$ 1.38 & 3.00 $\pm$ 1.02 & 7.08 $\pm$ 1.07 & 6.62 $\pm$ 1.07 & 3.31 $\pm$ 1.43 & 7.42 $\pm$ 1.06 \\

FreeTraj [refined] & 6.96 $\pm$ 1.12 & 4.85 $\pm$ 1.36 & 7.50 $\pm$ 0.86 & 6.24 $\pm$ 1.57 & 4.73 $\pm$ 1.29 & 6.84 $\pm$ 1.06 & $N/A$ & $N/A$ & $N/A$ \\

T2I & 4.12 $\pm$ 1.05 & 5.31 $\pm$ 1.08 & 3.83 $\pm$ 1.21 & 5.64 $\pm$ 1.40 & 4.73 $\pm$ 1.34 & 4.48 $\pm$ 1.00 & 2.63 $\pm$ 1.38 & 3.55 $\pm$ 1.48 & 3.00 $\pm$ 1.37 \\

\midrule

Ours & 4.54 $\pm$ 1.11 & 5.88 $\pm$ 1.01 & 5.62 $\pm$ 1.13 & 3.92 $\pm$ 1.23 & 7.35 $\pm$ 1.33 & 4.68 $\pm$ 1.10 & 5.46 $\pm$ 1.47 & 7.12 $\pm$ 1.24 & 4.54 $\pm$ 1.43 \\

Ours [refined] & 7.92 $\pm$ 0.98 & \textbf{7.54 $\pm$ 0.86} & \textbf{8.08 $\pm$ 0.92} & 8.68 $\pm$ 0.77 & \textbf{8.04 $\pm$ 1.05} & 7.72 $\pm$ 0.92 & \textbf{8.67 $\pm$ 0.69} & \textbf{8.88 $\pm$ 0.75} & 7.62 $\pm$ 1.02 \\

\bottomrule
\end{tabular}
}
\label{tab:user_study_table}
\end{table*}

\begin{figure*}[!t]
\centering
\includegraphics[width=\textwidth]{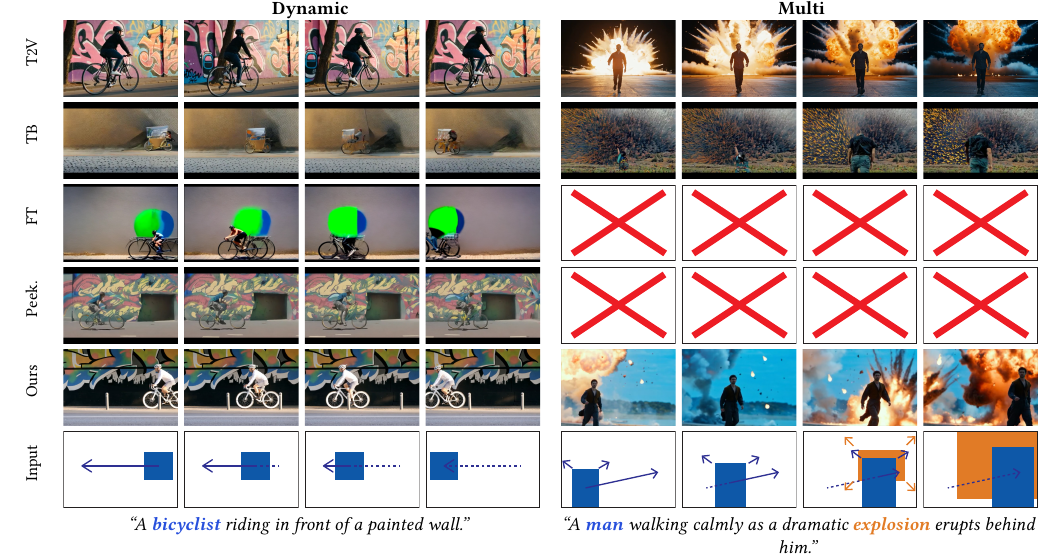}
\caption{
Comparison of our refined video (Ours) against competing methods.
Left: dynamic object layout. While T2V produces high-quality visuals, it does not follow the layout. Trailblazer (TB) and FreeTraj (FT) introduce bicycle artifacts, and Peekaboo (Peek.) shows weaker alignment. Right: multiple-object layout; empty frames indicate unsupported methods. Remaining methods fail to maintain spatial and temporal consistency: the explosion occurs prematurely, and the person does not follow the left-to-right trajectory specified by the blue box.
}
\label{fig:quality-dynamic-multi}
\Description{Comparison of outputs of different methods from 2 different datasets: Dynamic and Multi. The first example is of a bicyclist riding from right to left, the second example is of a man walking from left to right, and an explosion happens in the background.}
\end{figure*}

\Cref{fig:quality-dynamic-multi} demonstrate that our method outperforms prior approaches in terms of both layout alignment and overall video quality.
As stated earlier, during the guide generation stage we enforce strict layout adherence, which comes at the expense of visual fidelity, this quality is later recovered during the refinement stage. The strength of the optimization at the guide generation stage can be controlled by adjusting the learning rate.
\Cref{fig:optimization-strength,fig:t2i-abl-figure} illustrate the visual difference between the intermediate video guide and the final refined output.
This observation is further supported by the quantitative performance gap between stages reported in \Cref{tab:quantitative_table,tab:abl_layers,tab:abl_stage1_resolution}.

\edit{In our implementation of TrailBlazer on Wan 2.2, as well as prior benchmarks on earlier models, directly manipulating cross-attention maps leads to either artifacts or weak layout control, and these artifacts are often not recoverable during refinement (see \Cref{fig:quality-dynamic-multi}, supplemental). This further supports our claim that optimization-based guidance, together with a refinement stage, is necessary.}

Finally, we conduct a user study comparing our method with previous approaches \edit{on a subset of fifteen videos (five from each dataset)}. The study involves approximately sixty reviewers, including individuals with visual arts backgrounds as well as participants from non-visual backgrounds. In total, we collected 250 responses. Participants evaluated the generated videos along three dimensions: fidelity to the prompt, fidelity to the layout, and aesthetic quality. The aggregated results are shown in \Cref{tab:user_study_table}.
We observe that the straightforward Text-to-Video (T2V) method achieves strong visual quality (VQ) and prompt adherence, but performs poorly in adhering to layout guidance. Our approach significantly outperforms all baselines in layout adherence. In terms of VQ and prompt adherence, it performs competitively, surpassing all other methods except T2V. Notably, on certain datasets, our method also exceeds T2V in visual quality or prompt alignment.

\subsection{Ablation Analysis}

\begin{table*}[!htbp]
\centering
\caption{Ablation study on the contribution of different layers to layout adherence. Tested on Wan 2.2.
The results indicate that earlier layers perform better.
}
\resizebox{\textwidth}{!}{

\small

\begin{tabular}{c|c|ccccccc|ccccccc}
  \toprule
  \multirow{2}{*}{Range} & \multirow{2}{*}{\shortstack{\#\\layers}} &
  \multicolumn{7}{c|}{Guide} & \multicolumn{7}{c}{Refined} \\
  \cmidrule(lr){3-16}
   &  & IoU ($\uparrow$) & AP50 ($\uparrow$) & CD ($\downarrow$) & I. ($\uparrow$) & I-AP50 ($\uparrow$) & FVD ($\downarrow$) & ClipSim ($\uparrow$) &
   IoU ($\uparrow$) & AP50 ($\uparrow$) & CD ($\downarrow$) & I. ($\uparrow$) & I-AP50 ($\uparrow$) & FVD ($\downarrow$) & ClipSim ($\uparrow$)
   \\
  \midrule

  \multirow{2}{*}{0-10} & 3 & 33.82 & 17.35 & \textbf{8.31} & 86.15 & 90.65 & 173.48 & 29.69 & \textbf{27.84} & 12.61 & \textbf{11.18} & 74.00 & \textbf{78.66} &
   157.01 & 30.22 \\
   & 10 & \textbf{35.54} & \textbf{20.34} & 8.87 & \textbf{87.18} & \textbf{91.77} & 171.70 & 29.38 & 26.79 & 11.38 & 12.54 & 71.70 & 76.01 & 151.17 &
  29.60 \\
  \hline

  \multirow{2}{*}{10-20} & 3 & 31.94 & 19.73 & 10.52 & 82.04 & 89.18 & 161.22 & 29.15 & 24.83 & 11.61 & 13.61 & 69.72 & 75.84 & 150.15 & 30.33 \\
   & 10 & 25.49 & 10.88 & 14.12 & 72.42 & 75.65 & \textbf{156.44} & {29.74} & 19.64 & 6.24 & 16.66 & 58.56 & 62.59 & 145.02 & 29.97 \\
  \hline

  \multirow{2}{*}{20-30} & 3 & 16.56 & 1.91 & 21.42 & 51.48 & 53.81 & 162.91 & 29.45 & 15.70 & 2.18 & 21.06 & 48.33 & 50.43 & 150.00 & 30.02 \\
   & 10 & 29.95 & 18.67 & 13.51 & 69.12 & 76.36 & 167.67 & \textbf{29.91} & 24.75 & \textbf{14.03} & 14.86 & 61.91 & 65.35 & 153.51 & \textbf{30.34} \\
  \hline

  \multirow{2}{*}{30-40} & 3 & 15.21 & 2.55 & 22.23 & 50.03 & 49.83 & 156.91 & 29.50 & 15.20 & 3.58 & 21.79 & 47.68 & 46.15 & 147.73 & 29.89 \\
   & 10 & 16.22 & 1.22 & 21.89 & 50.04 & 52.01 & 165.54 & 29.63 & 17.01 & 2.98 & 19.50 & 51.14 & 54.32 & 150.82 & 30.21 \\
  \hline

  \multirow{3}{*}{0-20} & 3 & 27.84 & 8.88 & 11.46 & 79.36 & 86.67 & 169.42 & 29.39 & 22.14 & 6.63 & 13.88 & 65.48 & 71.46 & 146.24 & 29.92 \\
   & 10 & 32.78 & 18.95 & 9.58 & 86.80 & 90.68 & 167.28 & 29.40 & 25.70 & 12.45 & 12.67 & \textbf{74.31} & 78.21 & 152.81 & 29.97 \\
   & 20 & 33.00 & 17.59 & 9.40 & 85.45 & 90.65 & 169.42 & 28.97 & 26.35 & 12.41 & 11.86 & 73.01 & 76.52 & 150.46 & 29.91 \\
  \hline

  \multirow{3}{*}{20-40} & 3 & 16.61 & 2.79 & 21.60 & 53.20 & 55.54 & 158.88 & 29.51 & 16.22 & 2.74 & 20.66 & 50.26 & 52.84 & 150.36 & 30.02
  \\
   & 10 & 17.06 & 1.40 & 22.08 & 51.46 & 51.67 & 162.36 & 28.74 & 17.96 & 3.09 & 19.80 & 53.89 & 55.23 & \textbf{144.99} & 29.41 \\
   & 20 & 15.86 & 1.33 & 20.40 & 53.55 & 54.76 & 164.18 & 29.08 & 17.58 & 2.57 & 18.93 & 52.14 & 55.29 & 154.60 & 29.90 \\
  \hline

  \multirow{3}{*}{0-40} & 3 & 23.36 & 10.10 & 16.56 & 67.58 & 69.05 & 163.18 & 29.41 & 20.50 & 8.83 & 17.23 & 58.83 & 60.27 & 151.81 & 30.11 \\
   & 10 & 27.32 & 12.52 & 14.08 & 75.61 & 81.26 & 170.10 & 29.46 & 22.55 & 7.76 & 15.55 & 68.63 & 74.32 & 151.69 & 30.12 \\
   & 20 & 28.91 & 13.98 & 11.95 & 80.22 & 82.04 & 168.69 & 29.66 & 23.27 & 9.84 & 15.57 & 64.33 & 66.42 & 152.56 & 29.89 \\
  \bottomrule
  \end{tabular}

}
\label{tab:abl_layers}
\end{table*}

\begin{table*}[!htbp]
\centering
\caption{
Ablation study of resolution contributions to layout adherence. Resolution is denoted as $spatial$ $height$ $\times$ $temporal$ resolution.
\edit{Inference time of guide and refinement stages are compared to that of the backbone model.}
Focusing on the refined block (right) of the Wan model (top), the results indicate that it is generally better to generate the refined video at a higher resolution. At the highest resolution ($480 \times 81$) it is preferable in terms of trajectory accuracy (CD, I, I-AP50) to generate the respective guide at the lowest resolution (row 6), with negligible decrease in overlap accuracy (IoU, AP50).
}
\resizebox{\textwidth}{!}{
\setlength{\tabcolsep}{4pt}

\begin{tabular}{cc|cc|ccccccc|ccccccc}
% \hline
% & \multirow{2}{*}{\shortstack{Resolution\\(Guide/Refined)}} & \multirow{2}{*}{\shortstack{Inf. Time}}
% & \multicolumn{7}{c|}{Guide} & \multicolumn{7}{c}{Refined} \\
% \cline{4-17}
% &
% & & IoU ($\uparrow$) & AP50 ($\uparrow$) & CD ($\downarrow$) & I. ($\uparrow$) & I-AP50 ($\uparrow$) & FVD ($\downarrow$) & ClipSim ($\uparrow$) &
% IoU ($\uparrow$) & AP50 ($\uparrow$) & CD ($\downarrow$) & I. ($\uparrow$) & I-AP50 ($\uparrow$) & FVD ($\downarrow$) & ClipSim ($\uparrow$) \\
% \hline

\hline
& \multirow{2}{*}{\shortstack{Resolution\\(Guide/Refined)}} 
& \multicolumn{2}{c|}{\shortstack{Inference Time}} 
& \multicolumn{7}{c|}{Guide} 
& \multicolumn{7}{c}{Refined} \\
\cline{5-18}

&
& Guide/Refined & \edit{Backb.}
& IoU ($\uparrow$) & AP50 ($\uparrow$) & CD ($\downarrow$) & I. ($\uparrow$) & I-AP50 ($\uparrow$) & FVD ($\downarrow$) & ClipSim ($\uparrow$) &
IoU ($\uparrow$) & AP50 ($\uparrow$) & CD ($\downarrow$) & I. ($\uparrow$) & I-AP50 ($\uparrow$) & FVD ($\downarrow$) & ClipSim ($\uparrow$) \\
\hline

\multirow{10}{*}{\rotatebox{90}{Wan}}

& 160$\times$49/320$\times$49 & 00:36/01:10 & 01:13 & 37.01 & 24.15 & 7.82  & 86.00 & 92.48 & 174.01 & 28.94 & 22.67 & 10.31 & 15.23 & 57.75 & 62.72 & 156.44 & 30.60 \\
& 160$\times$49/320$\times$81 & 00:36/01:59 & 02:07 & 37.31 & 24.32 & 8.02  & 87.38 & 94.18 & \textbf{171.82} & 28.79 & 25.63 & 11.44 & 12.48 & 69.80 & 74.63 & \textbf{148.28} & 30.43 \\
& 160$\times$81/320$\times$81 & 00:50/01:59 & 02:07 & 37.54 & 24.69 & 8.63  & 87.22 & 91.13 & 181.94 & 29.09 & 24.90 & 9.42  & 13.49 & 69.85 & 72.49 & 152.32 & 30.38 \\
\cline{2-18}

& 320$\times$49/320$\times$81 & 01:55/01:59 & 02:07 & 46.74 & 40.75 & 5.73  & 89.49 & 94.59 & 185.68 & 27.48 & 28.20 & 12.72 & 11.30 & 71.27 & 75.64 & 158.87 & 30.61 \\
& 320$\times$81/320$\times$81 & 03:21/01:59 & 02:07 & 48.88 & 46.40 & \textbf{5.34} & \textbf{90.34} & \textbf{96.32} & 181.73 & 26.66 & 27.62 & 14.55 & 10.77 & 71.41 & 74.12 & 163.21 & 30.35 \\
\cline{2-18}

& 160$\times$49/480$\times$81 & 00:36/05:58 & 06:20 & 35.84 & 21.57 & 8.32  & 86.31 & 92.96 & 172.46 & 29.15 & 29.50 & 14.90 & \textbf{9.67} & \textbf{78.04} & \textbf{83.33} & 148.46 & 29.90 \\
& 160$\times$81/480$\times$81 & 00:50/05:58 & 06:20 & 35.22 & 21.40 & 10.05 & 83.53 & 87.55 & 185.56 & 28.52 & 29.62 & 18.23 & 11.09 & 75.88 & 78.60 & 156.97 & 28.96 \\
& 320$\times$49/480$\times$81  & 01:55/05:58 & 06:20 & 45.69 & 39.01 & 6.26  & 89.38 & 94.15 & 183.60 & 27.91 & 31.67 & 17.98 & 9.85  & 74.66 & 77.68 & 164.40 & 29.77 \\
& 320$\times$81/480$\times$81 & 03:21/05:58 & 06:20 & 48.08 & 45.93 & 6.23  & 88.51 & 93.46 & 183.80 & 27.27 & 27.45 & 14.82 & 12.39 & 67.62 & 69.86 & 161.05 & 30.14 \\
& 480$\times$81/480$\times$81 & 10:24/05:58 & 06:20 & \textbf{51.36} & \textbf{58.25} & 6.03 & 88.57 & 95.06 & 192.49 & 26.33 & 31.96 & 17.41 & 10.39 & 72.06 & 77.63 & 176.14 & 29.96 \\
\hline

\multirow{3}{*}{\rotatebox{90}{LTX}}

& 128$\times$121/256$\times$121  & 00:26/00:01 & 00:08 & 20.57 & 7.25  & 21.66 & 46.55 & 48.95 & 176.92 & \textbf{31.52} & 19.87 & 7.22  & 19.87 & 44.47 & 47.23 & 162.07 & \textbf{32.17} \\
& 256$\times$121/512$\times$121 & 00:34/00:02 & 00:10  & 31.63 & 21.10 & 13.14 & 62.27 & 69.34 & 178.15 & 31.35 & 28.57 & 14.46 & 13.72 & 61.54 & 67.12 & 162.73 & 31.24 \\
& 384$\times$121/768$\times$121 & 01:06/00:04 & 00:22 & 34.30 & 21.50 & 13.08 & 71.60 & 76.17 & 173.00 & 30.08 & \textbf{35.60} & \textbf{24.19} & 9.70 & 74.92 & 80.58 & 164.22 & 30.84

\end{tabular}

}
\label{tab:abl_stage1_resolution}
\end{table*}

We conduct a series of ablation studies to better understand the design choices in our method. In particular, we analyze (i) which layers are most influential for layout control generation, (ii) how the number of optimized layers impacts output quality, and (iii) how the resolution used at different stages affects the trade-off between inference time and performance.

In \Cref{tab:abl_layers} we study which layers to select and how many are needed. Instead of fixing a single subset, we randomly sample a subset from the chosen layer range at each diffusion timestep. This strategy benefits from representations across the full layer set while reducing computational cost for high-resolution inputs. Our results indicate that early layers are especially beneficial.  Specifically, during the guide generation stage (left columns block), we observe higher IoU and AP50 and lower CD for early layer ranges (0–10, 10–20, 0–20) compared to later ranges (20–30, 30–40, 20–40). This advantage consistently propagates to the refinement stage (right columns block). 

\Cref{tab:abl_stage1_resolution} shows the benefit of generating the refined video at a higher resolution, but inference time is greatly reduced when the guide is generated at lower resolutions with only a slight drop in performance. Due to its high compression rate, LTX achieves significantly faster inference times. Wan attains the strongest performance during the guide stage, while LTX achieves higher IoU and AP50 scores after refinement, along with comparable performance on trajectory-related metrics (CD, Intersection, and I-AP50).

For all results in the paper, we adopt the setting with three random layers selected from layers $0-10$ (see \Cref{tab:abl_layers}, first row).
We use the lowest guide resolution ($160 \times 49$) and the highest refined resolutions ($480 \times 81$, $768 \times 121$) from \Cref{tab:abl_stage1_resolution} (rows 6 and 13), as they yield the best overlap and trajectory accuracy in the refined videos.
Due to paper format limitations, we refer the reader to the supplemental for video demonstrations and further experiments.

\subsection{Additional Applications}

\begin{figure*}[htbp]
\centering
\includegraphics[width=\textwidth]{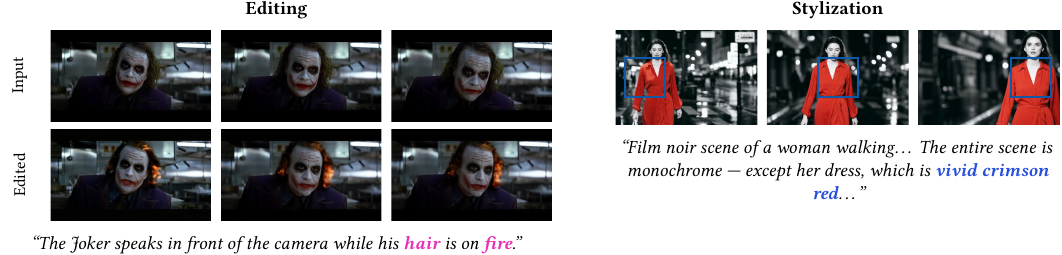}
\caption{
Additional applications of our method.
Left: video editing, where the input video is transformed according to the text prompt.
Right: stylization, where only selected semantic elements retain color while the rest of the scene remains monochrome.
}
\label{fig:additional-app}
\Description{
Side-by-side comparison of two applications. Left: a Joker video edited so his hair appears on fire.
Right: a film noir stylization where only a woman's red dress remains colored.
}
\end{figure*}

We show more uses of our method in \Cref{fig:additional-app}. 
By generating the prior from an existing video, we can edit its attributes. Applying optimization to style-related parts in the prompt, produces different styles in the same video.
More examples and details are in the supplemental.

% FIGURE PAGE 2
\begin{figure*}[!t]
\centering
\includegraphics[width=\textwidth]{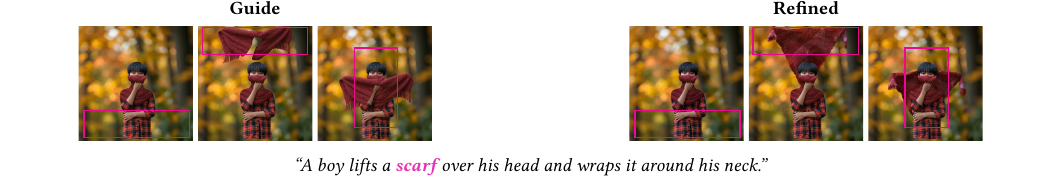}
\caption{
Comparison between the video guide generated using an ensemble of Text-to-Image methods (left) and the refined video produced by a pre-trained video model (right).
The guide fails to capture the interaction between the boy and the scarf: the hands remain static instead of grasping and moving the scarf, resulting in motion that lacks coordination between the objects.
}
\label{fig:t2i-boy-scarf}
\Description{Side-by-side comparison of guide and refined video frames of a boy lifting a scarf over his head.}
\end{figure*}

\section{Limitations}

Despite its strong performance, the approach has several limitations. It does not explicitly separate camera motion from object motion: bounding boxes define where content should appear over time, but they do not specify whether changes should arise from object movement or from camera transformations, which can lead to ambiguous or unintended behavior.

Performance may decrease in complex situations, such as scenes with detailed motion, multiple interacting objects, or crowded scene layouts. Although we demonstrate results in such settings, they remain challenging and leave room for further improvement.
An important direction for future work will be to address these challenges, which may lead to more robust layout-guided video generation.

\section{Conclusion}

We present a training-free approach for layout-guided text-to-video generation that enables explicit control over spatial composition and object motion while preserving the generative strength of modern video diffusion models. 

Our results show that minimal intervention at inference time is sufficient to steer pre-trained models toward user-specified spatiotemporal layouts without compromising visual fidelity or requiring additional training. Through extensive quantitative and qualitative evaluation, we demonstrate consistent improvements over prior methods across static, dynamic, and multi-object scenarios.

% \clearpage

% DO NOT INCLUDE ACKNOWLEDGMENTS IN AN ANONYMOUS SUBMISSION

% \newpage
\clearpage
\twocolumn
% Bibliography
\bibliographystyle{ACM-Reference-Format}
\bibliography{main}

% Appendix
\appendix

\end{document}